\documentclass[sigchi-a, nonacm=true]{acmart}

\def\BibTeX{{\rm B\kern-.05em{\sc i\kern-.025em b}\kern-.08emT\kern-.1667em\lower.7ex\hbox{E}\kern-.125emX}}
    
\copyrightyear{2019}
\acmYear{2019}
\setcopyright{acmlicensed}
\acmConference[CHI '19 Workshop: Challenges of Working on Social Robots...]{CHI 2019 Workshop: The Challenges of Working on Social Robots that Collaborate with People}{May 04--09, 2019}{Glasgow, UK}
\acmBooktitle{CHI 2019 Workshop: The Challenges of Working on Social Robots that Collaborate with People, May 04--09, 2019, Glasgow, UK}
\acmPrice{15.00}
\acmDOI{10.1145/1122445.1122456}
\acmISBN{978-1-4503-9999-9/18/06}

\usepackage[utf8]{inputenc}
\usepackage[english]{babel}
\usepackage{natbib}

\usepackage[bottom]{footmisc}
\begin{document}

%
\title{Challenges in Collaborative HRI for Remote Robot Teams}

%
\author{Helen Hastie}
\email{H.Hastie@hw.ac.uk}

\author{David A. Robb}
\email{D.A.Robb@hw.ac.uk}

\author{Jos\'{e} Lopes}
\email{JD.Lopes@hw.ac.uk}

\author{Muneeb Ahmad}
\email{M.Ahmad@hw.ac.uk}

\author{Pierre Le Bras}
\email{P.Le_Bras@hw.ac.uk}

\affiliation{%
}

\author{Xingkun Liu}
\email{X.Liu@hw.ac.uk}

\author{Ronald P. A. Petrick}
\email{R.Petrick@hw.ac.uk}

\author{Katrin Lohan}
\email{K.Lohan@hw.ac.uk}

\author{Mike J. Chantler}
\email{M.J.Chantler@hw.ac.uk}

\affiliation{%
  \institution{Heriot-Watt University}
  \streetaddress{Riccarton}
  \city{Edinburgh}
  \country{UK}
  \postcode{EH14 4AS}
}
\renewcommand{\shortauthors}{Hastie et al.}


%
\begin{abstract}
Collaboration between human supervisors and remote teams of robots is highly challenging, particularly in high-stakes, distant, hazardous locations, such as off-shore energy platforms. In order for these teams of robots to truly be beneficial, they need to be trusted to operate autonomously, performing tasks such as inspection and emergency response, thus reducing the number of personnel placed in harm's way.  As remote robots are generally trusted less than robots in close-proximity, we present a solution to instil trust in the operator through a `mediator robot' that can exhibit social skills, alongside sophisticated visualisation techniques.  In this position paper, we present general challenges and then take a closer look at one challenge in particular, discussing an initial study, which investigates the relationship between the level of control the supervisor hands over to the mediator robot and how this affects their trust. We show that the supervisor is more likely to have higher trust overall if their initial experience involves handing over control of the emergency situation to the robotic assistant. We discuss this result, here, as well as other challenges and interaction techniques for human-robot collaboration.




\end{abstract}

%
%

\begin{marginfigure}
  \centering
  \includegraphics[width=1\linewidth]{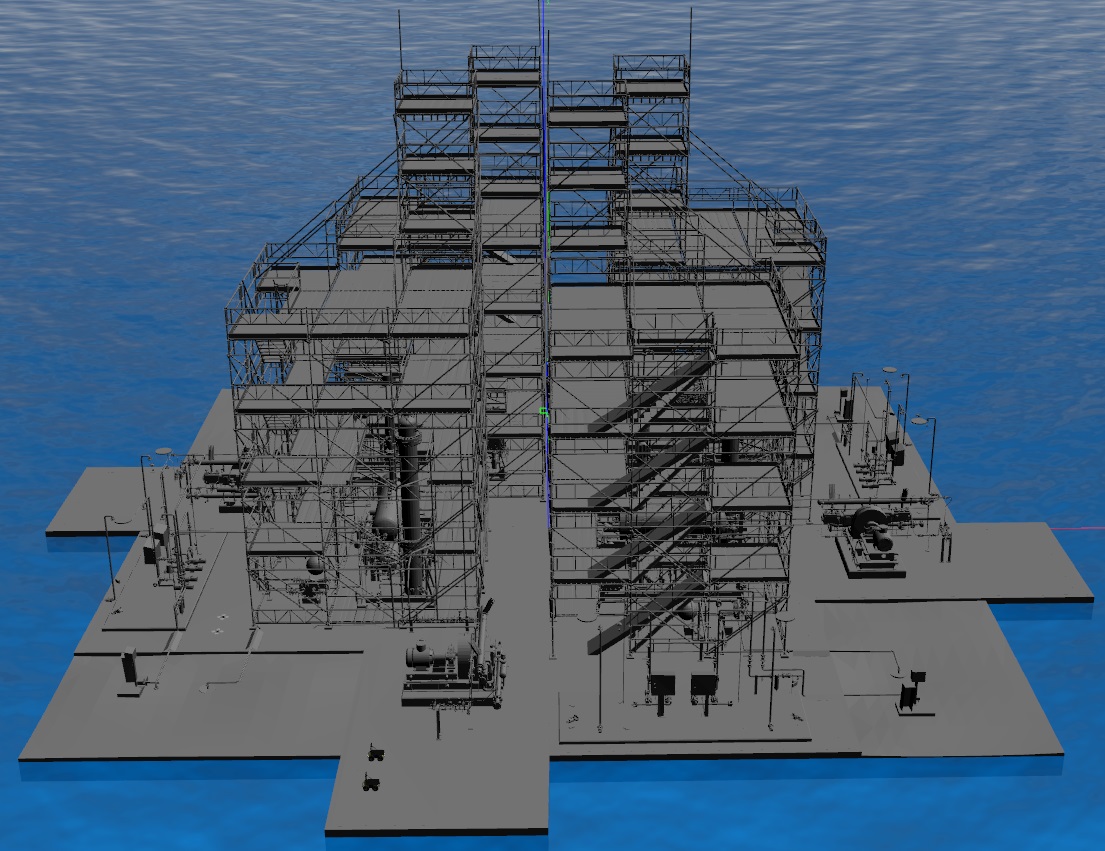}
  \caption{Digital Twin of an offshore platform for collaboration with remote robots. Implemented in Gazebo and ROS, it simulates the restrictions to movement in a complex structure that robot teams can be expected to encounter and their supervisors will need to manage.}
  \Description{Simulation environment.}
  \label{fig:gezeborig}
\end{marginfigure}

\begin{marginfigure}
  \centering
  \includegraphics[width=\linewidth]{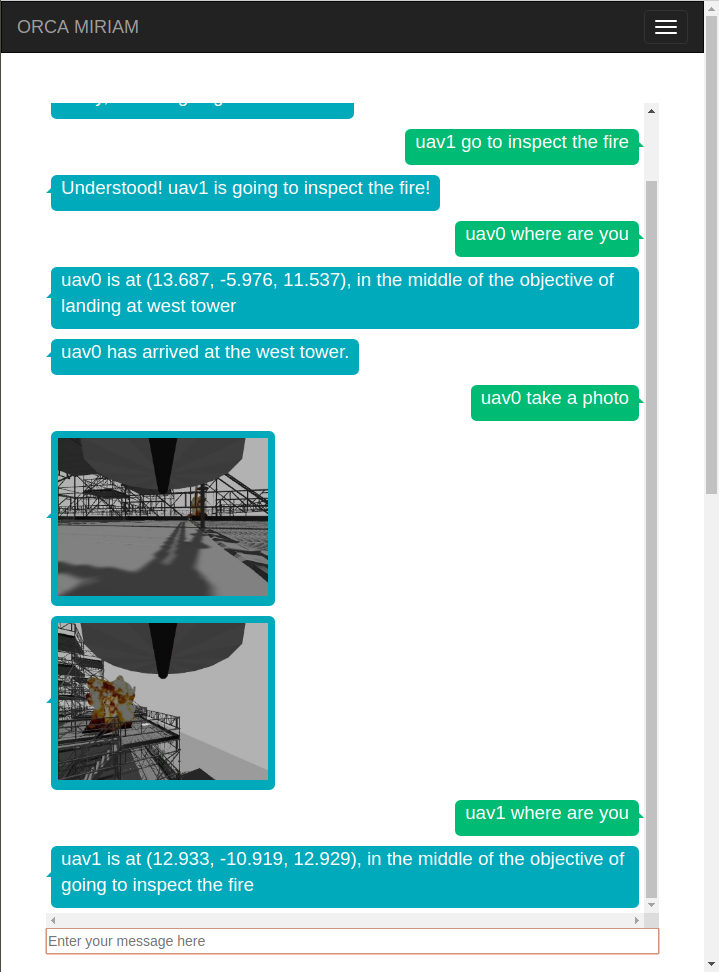}
  \caption{Natural language interaction in simulation between the operator and remote robots.}
  \Description{Simulation environment.}
  \label{fig:miriam}
\end{marginfigure}



%

%
\maketitle


\begin{sidebar}
      \caption{Main challenges for collaboration with remote robots.}
      \label{sidebar:main_challenges}
      
      \begin{enumerate} 
\item {\bf Planning in human-robot teams}: Generation of coherent plans for safe and efficient human-robot teamed activity at the remote location; communication of the plans to the team supervisors, through language/visualisation.

\item {\bf Executing/monitoring a task by Human-Robot teams}:  Situation awareness providing monitoring information during operations as the plans unfold in action, through human-robot interaction or sophisticated visualisation techniques. 

\item {\bf Adaptivity in Human-Robot teams}: Joint problem solving and re-planning in the light of unforeseen events, change in personnel or conditions, supervisor decisions and changing priorities. If shared control is necessary, when and how should it be offered.

\end{enumerate}

\end{sidebar}

\section{Introduction}
Robots and autonomous systems are increasingly being used to perform tasks 
where it is either impossible or dangerous for humans to go, such as deep underwater or other remote locations. These systems may be deployed, for example, in the energy sector, search and rescue or defence
\cite{Kwon12, Nagatani13, Shukla16, Trevelyan2016, Wong17}.
Our work on the ORCA Hub project \cite{orcashortform} envisages that teams of remote robots will be required to inspect and service remote off-shore energy installations, such as oil and gas platforms and wind farms. A 3D simulation environment is shown in Figure \ref{fig:gezeborig}, which is used to rehearse scenarios and evaluate team collaboration methods offshore \cite{digitaltwin19shortform,Chen17shortform}. 
The goal is to instil these robots with high levels of autonomy and have them operate in self-organising teams, as well as in mixed teams with humans and robots collaborating effectively to get the job done. Increasing levels of autonomy require a new approach to communication, especially in dispersed groups with human users of varying expertise, mental models and goals. Traditionally, robots have been remotely operated in these types of scenarios and this new dynamic requires a shift from user-manipulation of a tool to a collaborative relationship between human and machine \cite{toosl2teammatesshortform,SchaeferTrustTeamshortform}.

Successful team collaboration can depend on a variety of complex, interwoven factors. Prior work has examined how 
trust affects team collaboration \cite{Castelfranchi2010shortform,Langfred2004shortform,Hancock2011trustshortform}, 
which is particularly challenging with robots that are out-of-sight, as studies have shown that remote robots instil less trust than local robots \cite{Bainbridge2008shortform}. Other factors include communication and interaction, which can help with establishing trust, but also information sharing \cite{Mesmer-magnus2009shortform}, establishing common ground \cite{Stubbs2007shortform}, mental models \cite{Garcia18shortform} and team cognition \cite{Cooke2015shortform}.  
Communication of intent is critical both on the human side \cite{AllenPerraultshortform} and the robot \cite{Schaefer_chen_etal2017shortform}, with this 
closely linked to situation awareness and monitoring \cite{Robb2018shortform}. Finally, the ability to take control when and if needed has a non-trivial relationship with trust and team collaboration \cite{Castelfranchi2010shortform} and is discussed below in reference to an initial study reported here. 

Overarching themes across these factors include anthropromorphism, transparency of the system (e.g. through explanations), adapting in the face of errors, as well as, user variation and cognitive load.  We discuss these various challenges for collaboration in terms of three stages of missions (as summarised in Sidebar \ref{sidebar:main_challenges}): planning, execution/monitoring and adapting to unseen events.  We set out results and discussions stemming from our study and provide general discussion on future challenges in this area.

\section{Planning in Human-Robot Teams}

As systems become more autonomous, control diverts away from human operators to automated decision-making mechanisms such as planning. Here, the user sets a high-level goal (e.g. `inspect the north tower') and then the subsequent system's decisions would no longer be determined directly by the human operator. The operator, nevertheless, plays a key role in the process and may be required to oversee overall operations, adjust parameters within the control mechanisms, modify non-optimal behaviour, or take over some degree of control from the agent during task execution.   Therefore, in order for humans to interact collaboratively with robots, a number of planning challenges (Sidebar \ref{sidebar:main_challenges}, item 1) must be addressed due to the presence of humans in the planning loop \cite{Kambhampati-Talamadupula:2015shortform}. These include, ensuring robot actions do not conflict with human activities for safety reasons (human-aware planning), enabling robots to work together effectively with other robots and humans for efficient task completion (collaborative multiagent planning), and providing decision support for humans to inspect, evaluate, and revise generated plans (mixed-initiative planning).
\begin{marginfigure}
  \centering
    \includegraphics[width=0.55\linewidth]{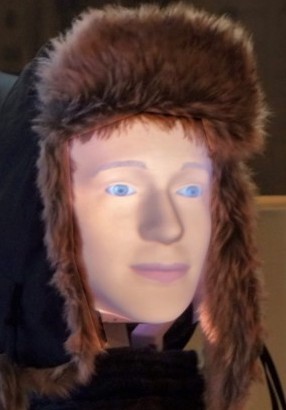}
  \caption{The ORCA social robot acting as a `mediator robot' between the humans and the remote robots. This mediator robot was embodied in a Furhat robot, which is able to exhibit social cues through facial expressions and neck movement. Image supplied by Furhat Robotics.}
  \Description{The Furhat robot used to embody a robot assistant. Image supplied by Furhat Robotics.}
  \label{fig:furhat}
\end{marginfigure}
Factors affecting plans include constraints such as battery life and robot capabilities, along with the actual task objectives, e.g., inspecting a particular item of infrastructure on the platform. On ORCA, these are being addressed by means of a temporal planning approach  \cite{Benton-etal:2012,Crosby-Petrick:2014shortform}, where the actions and goals of multiple robots (and possibly humans), along with their interdependencies, are considered as part of the plan generation process.


\begin{marginfigure}
  \centering
  \includegraphics[width=\linewidth]{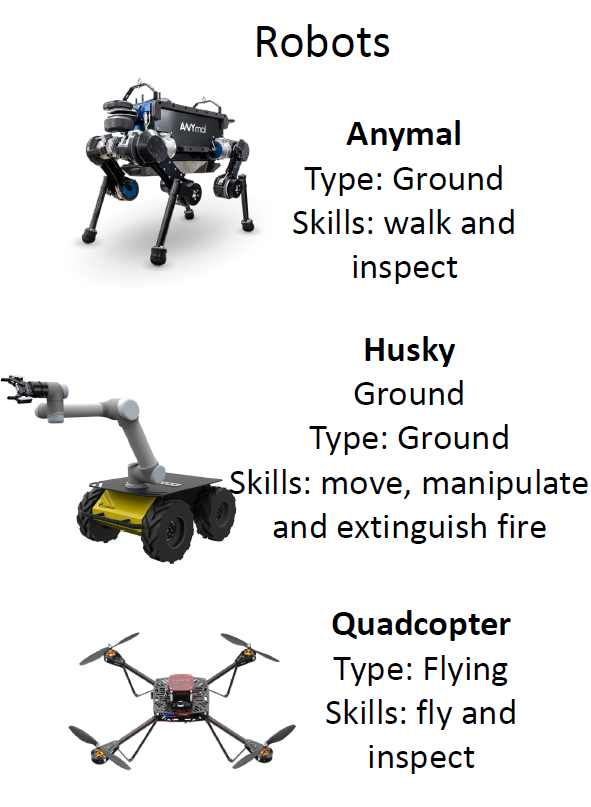}
  \caption{Remote robot team that that the mediator robot can control.}
  \Description{Remote robots that the Furhat robot assistant controls in the a simulation study.}
  \label{fig:robots}
\end{marginfigure}

\section{Executing/Monitoring a Task in Human-Robot Teams}
Understanding the intent and how the system makes decisions is critical for situation awareness and team collaboration \cite{Schaefer_chen_etal2017shortform}. Thus the human component of the team needs to monitor and understand both {\it what} the system is doing and also {\it why}. One study showed that a conversational agent (MIRIAM), alongside a graphical display of system status, can improve operator situation awareness \cite{Robb2018shortform}.  This conversational agent was able to provide natural language alerts and on-demand situation awareness in terms of mission progress, whilst also being able to provide explanations of underwater autonomous system behaviour in terms of current high level goals, thus improving the mental model of the operator \cite{Garcia18shortform}. Example interaction includes: ``Why are you spiralling up?", the response being, ``....to do a GPS fix". 

Plan execution must be monitored to assess plan success and failure. On ORCA, this will be done with a plan explanation interface providing information to humans about plan decisions and outcomes. 
The role of such an interface comprises two parts. Firstly, it should inform the supervisor about the details of any newly generated robot plans, specifically: general trend of the plan course, potential risks of conflicts between agents (humans and robots), and estimation of the robots' states (e.g. battery life) throughout the plan. Secondly, it should provide means for the supervisor to interact with the planning system and decide on the best course of action based on their expertise, specifically a choice between a set of alternative plans, modification of a plan, approval of only part of a plan, and re-definition of goals and heuristics. These forms of interaction have been previously identified as desirable goals for temporal planning interfaces in general \cite{Freedman2018shortform}.






\section{Adaptivity in Human-Robot Teams}
The ability for teams to be flexible and act quickly to a changing world/environment is key for an effective team. 
The interaction between the supervising human operators and the system for dealing with such unforeseen circumstances will be complex. The supervisors will be called on to make decisions on the appropriateness of plans in a challenging, fast-moving environment and be required to be accountable for the outcomes of the activities.  As the supervisors will be remote from the action, they will need to collaborate with the system in choosing and approving plans and monitoring, as described above, but also react to events as they occur. For this collaboration to work, the supervisors will need to trust the abilities of the system to achieve shared goals \cite{Chen:2018:PTH:3171221.3171264shortform}.

\begin{marginfigure}
  \centering
  \includegraphics[width=\linewidth]{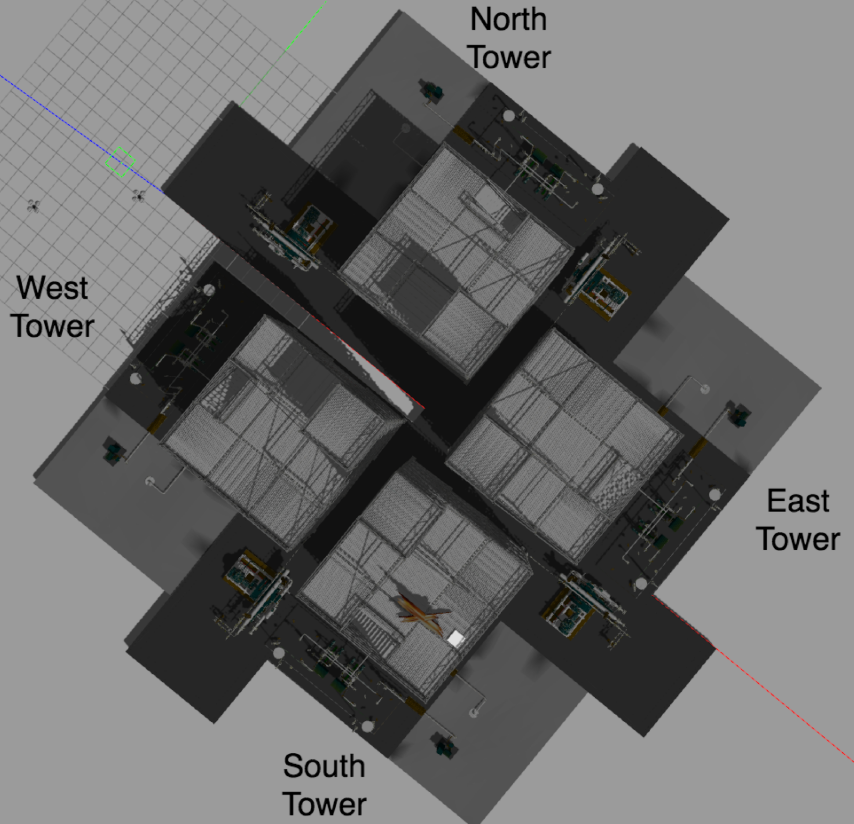}
  \caption{The tower plan shown to participants in the study described in the STUDY section.  }
  \Description{The tower plan shown to participants in the study described in the STUDY section.}
  \label{fig:towerplan}
\end{marginfigure}

\begin{sidebar}
      \caption{Example dialogues from experimental sessions where participants had varying levels of control.}
      \label{sidebar:example_dialogs}
\textit{\textbf{High User Control:}}\\
         \textbf{S}: Which robot should we send to put out the fire?\\  
         \textbf{U}: The Husky please\\ 
         \textbf{S}: Moving Husky to processing module East Tower\\
         \textbf{U}: Can you send a Quadcopter to the North Tower as well?\\
        \textbf{S}: Okay.\\
        
       \textit{\textbf{Low User Control:}}\\
         \textbf{S}: I'll check which robots are available to resolve the fire. \\  
         \textbf{S}: Husky 1 is available. \\
         \textbf{S}: Should Husky 1 be sent to processing module East Tower?  \\
         \textbf{U}: Yes. \\
        \textbf{S}: Okay. \\
      
\end{sidebar}

Part of adapting to an unforeseen situation is {\it if and when} to take control of the situation either by replanning or taking direct control through remote operation. From the outset, one might presume that the more one controls the system the less one trusts it and  similarly, the more one trusts the system, the less one needs to control it.  However, the relationship between trust and control is a complex one, as discussed in 
\cite{Castelfranchi2010shortform}, where they posit that, 
when taking into account the broader picture, control and trust can be complementary (what they call `broad trust'). This topic of control and trust is particularly important for distributed systems with varying goals, knowledge and competencies as discussed below with respect to our study. 

Finally, one of the main challenges of team work is adapting to the various roles of the users and the varying knowledge/mental models of the individuals in the team (humans and robots). The team can be constantly changing due to, for example, personnel change across shifts. This can be further compounded by individual differences such as  socio-cultural differences and propensity to trust \cite{PropensityScale}. 
Finally, if we are asking operators to manage multiple robots and robot-human teams that are remote and in high-stakes environments, there is a risk of cognitive overload of the operator \cite{lopes2018symptomsshortform}. Sensing and adapting interaction to both the user state and world state is key to effective collaboration and ensuring the human is not overloaded.

\section{A Social Robot As a Mediator}
Previous approaches have mostly been in the style of traditional user interfaces \cite{Schaefer_chen_etal2017shortform}.
Here, we present an approach that uses a conversational agent to allow for natural and fluid communication. This can be achieved through text chat, similar to group chat systems for human-teams. An ORCA prototype of such a group chat is illustrated in Figure \ref{fig:miriam}, whereby the supervisors can directly communicate with the robots to obtain status and set high-level goals (e.g. "UAV1 inspect the north tower"). An alternative is to employ a conversational assistant, such as MIRIAM, which is able to give an overview of multiple vehicles on a mission though natural language text or voice \cite{hastie2017miriamdemoshortform}. Voice as an alternative to text allows for hands-free access 
through mobile phones or radio \cite{daniels2004establishingshortform}. 

Using a conversational social robot, such as the Furhat (see Figure \ref{fig:furhat}), further allows users to interact in a social manner, which can assist in increasing the supervisor's trust levels by subtly and naturally indicating thoughtfulness,  certainty, confidence as well as providing cues to take over control through 
face-to-face conversation  \cite{foster2007enhancingshortform,loth2018confidenceshortform}.
\begin{sidebar}
      \caption{Study set-up.}
     \label{sidebar:set-up}
The experiment was planned as a repeated measures experiment with two conditions, Low and High Control as the independent variable and measuring User Trust as the dependent variable.  25 participants were recruited from our campus and offered a reward of the chance to win a shopping voucher. The order of these interactions was balanced such that half the participants (assigned randomly) did Low Control first and half High Control first.  
Participants were told that they would interact with a robot assistant to resolve an emergency on an oil rig using a team of robots. They were up against a time limit of 3 minutes. If they did not resolve the emergency in time, then the rig would need to be evacuated. They were given details of their robot team and the rig on a sheet of paper (Figures \ref{fig:robots} and \ref{fig:towerplan}). 

\end{sidebar}
Incorporating a social robot into the team of humans and remote robots is motivated by findings such as that reported in \cite{waytz2014mindshortform}, whereby 
drivers of autonomous vehicles trusted their vehicles' autonomous features more when they were viewed with  a level of anthropromorphism, i.e. had human-features such as names and a voice. 
In addition,  \cite{d566fe253b974acdad3cd3b865ac26c2} show that there is a clear difference in how participants trust robots that support them, depending on the robots' appearance. Having a social robot as a mediator allows for an element of anthropromorphism, where it is not possible to implement them in the physical remote robot team, due to their distance and number. In these types of scenarios, there could be a large number of unmanned air vehicles (UAVs), land/marine surface vehicles and underwater vehicles and it would be infeasible for a human to remember all their names or bond with them.  In addition, these robots are designed to perform a function without a view as to how they look or interact with humans (e.g. UAVs with camera).

Compounding this is that in remote environments where robots are out-of-sight and communication may be intermittent, trust is harder to establish \cite{Bainbridge2008shortform}. There exists much uncertainty, for example, when supervising and monitoring underwater robots where there is only intermittent communication, and there may be more than one possible explanation of what they are doing \cite{Garcia18shortform}. Language can be used to portray this uncertainty \cite{dimitrauncertaintyshortform} and social robots could help in this regard, as they can convey social cues for uncertainty visually through bodily or facial features \cite{loth2018confidenceshortform,lohan2016enrichingshortform,fischer2012levelsshortform,vollmer2009peopleshortform}.  Whether to give the user only information that the system is 100\% certain of, or give them all the options, has been the topic of research  \cite{Garcia18shortform} and is clearly dependent on the domain and the level of critical safety.

Finally, social robots could help with explaining plans. Previous studies have shown that providing "thoughtful interaction", for instance where robots ponder plans, can increase the confidence and trust of the user \cite{desai2012effectsshortform}. With respect to all of the above, care needs to be taken when choosing the embodiment and level of human-likeness, so as not to unnerve the human members of the team \cite{mori1970uncannyshortform}.

\section{Study}
\label{sec:Study}

Here, we describe an initial study that addresses one of the aspects of team collaboration discussed above, namely the amount of control the human team-mate should have and how this affects trust (see Sidebar \ref{sidebar:set-up}). 
We investigate, here, control in terms of defining high-level goals such as "send the husky robot to inspect the north tower" (rather than low level planning) and whether these should be left to the mediator robot, who follows a standard emergency response protocol, or whether the human supervisor feels more comfortable controlling the situation themselves. The robot assistant interaction was implemented using the Wizard-of-Oz technique combined with a Finite State Automaton to follow a sequence of steps in order to complete the mission \cite{lopes_hrilbr19shortform}.




 The participants' task was to collaborate in order to resolve an emergency on the rig in three steps: 1) identify and precisely locate the emergency, 2) resolve the emergency and 3) assess the damage caused. Participants undertook two interactions (two conditions), where they were given differing levels of control:  a low level of control involving agreeing to high-level actions proposed by the robot assistant (Sidebar \ref{sidebar:example_dialogs}, bottom); and a high level of control involving choosing the robots to be deployed and where (Sidebar \ref{sidebar:example_dialogs}, top). After each interaction, user trust was measured using the 14-item Schaefer trust questionnaire \cite{schaefer2013perceptionshortform}. 

\begin{sidebar}
      \caption{Summary of study results.}
     \label{sidebar:results_summary}
         Overall, we found no significant difference in our data between the two conditions (High and Low Control, see examples in Sidebar \ref{sidebar:example_dialogs}).
\textit{\textbf{However, interesting significant differences were found}} when we analysed the results as being for two groups depending on whether they experienced Low Control first (LC-first), or High Control first (HC-first). This splitting follows the idea that there is some bias effect from the first version of the system participants interacted with that would be reflected in the interaction data during, and in the questionnaires after, the second interaction.

 To test the questionnaire answers for significance, we performed a Wilcoxon-Mann test and found a significant effect for one component of the trust questionnaire \cite{schaefer2013perceptionshortform}, specifically `\textit{percentage of the time that the robot-mediator will have errors}' in the HC-first condition ($Mdn = 20$) and LC-first condition ($Mdn = 10$), $W=429.5$, $p<0.05$. This indicates that those with LC initially trusted the system more to have fewer errors (lower score indicates higher trust in terms of likelihood to have errors). 
\end{sidebar}

This study provides initial experimental results (see Sidebar \ref{sidebar:results_summary}) that subjects were more likely to have higher trust overall if their initial experience was having a low level of control. Therefore, at least initially, handing-off control results in higher trust in terms of how much the participants trust the robot not to make errors. This aligns with the traditional thinking that trust and control are antagonistic of each other, with one compensating for the other. For our non-expert subjects, this makes sense whereby the robot, in essence, shows the human what types of actions result in a successful mission, enabling the user to take over in the second stage with confidence.  

\vspace{-0.25cm}
\section{Discussion and future work}

The results of our study may be different with expert users, where they may prefer continuous high levels of control, where trust and control would be complementary.  Indeed, a survey of HRI trust literature \cite{Hancock2011trustshortform} does identify user expertise as affecting trust. This leads us to suggest that initial levels of control ceded to the user by the assistant should be varied according to user role and expertise. Hancock et al. \cite{Hancock2011trustshortform} also indicate anthropomorphism as a factor and our experience in the study suggests that we could, in future, add more expressiveness to our assistant. We did use eye blink behaviours in the Furhat robot but intend using head/neck movements and more expressive intonation to engage users in future. 

One interesting area is what happens if an error occurs and how the mediator robot might use social functions to regain the trust.  Mirnig et al. \cite{mirnig2017errshortform} have explored human reactions on robot error/faulty behaviour and found that, to understand the situation, the decoding of the human's social signals is required.  We see a possibility of using gaps in task-relevant conversation to establish empathy \cite{Darling_Breazeal_etal2015shortform} and build trust. In our study, there were substantial pauses in the dialogue while user and assistant waited for the remote robots to move and act on instructions. This would naturally depend on the scenario, with emergency response having less room for social interaction, as compared to routine work such as inspections, when social interactions might help maintain supervisor focus. In addition, interaction that indicates thoughtfulness and even expressions of remorse for errors,  may help with trust \cite{waytz2014mindshortform} and therefore, team collaboration. This still needs to be investigated, in particular in the context of time critical scenarios as in the ORCA domain.

\begin{acks}
We acknowledge funding and support from the EPSRC ORCA Hub (EP/R026173/1, 2017-2021) and consortium partners.
\end{acks}
%
\bibliographystyle{ACM-Reference-Format}


\bibliography{HERE-sigchi-a}

\end{document}